\pdfoutput=1

\documentclass[11pt]{article}

\usepackage[preprint]{acl}

\usepackage{times}
\usepackage{latexsym}

\usepackage[T1]{fontenc}

\usepackage[utf8]{inputenc}

\usepackage{microtype}

\usepackage{inconsolata}

\usepackage{graphicx}
\usepackage{tikz}
\usepackage{pgfplots}
\usepackage{pgfplotstable}
\usepackage{amsmath}
\usepackage{subcaption}
\usepackage{amssymb}
\usepackage{booktabs}
\usepackage{multirow}
\usepackage{makecell}

\usetikzlibrary{positioning, shapes, calc, fit, arrows.meta, decorations.pathreplacing}

\newcommand\defeq{\mathrel{\stackrel{\makebox[0pt]{\mbox{\normalfont\scriptsize def}}}{:=\,}}}

\newcommand{\CommentedText}[1]{}

\title{LoRA-BAM: Input Filtering for Fine-tuned LLMs via Boxed Abstraction Monitors over LoRA Layers}

\author{
  \textbf{Changshun Wu\textsuperscript{1}},
  \textbf{Tianyi Duan\textsuperscript{2}},
  \textbf{Saddek Bensalem\textsuperscript{3}},
  \textbf{Chih-Hong Cheng\textsuperscript{2,4}}\thanks{The first two authors contributed to the work equally.}
  \\
  \textsuperscript{1}Universit\'e Grenoble Alpes, Grenoble, France\\
  \textsuperscript{2}Chalmers University of Technology, Gothenburg, Sweden\\
  \textsuperscript{3}CSX-AI, Grenoble, France\\
  \textsuperscript{4}University of Gothenburg, Gothenburg, Sweden
}

\begin{document}
\maketitle
\begin{abstract}

Fine-tuning large language models (LLMs) improves performance on domain-specific tasks but can lead to overfitting, making them unreliable on out-of-distribution (OoD) queries. We propose LoRA-BAM - a method that adds OoD detection monitors to the LoRA layer using boxed abstraction to filter questions beyond the model’s competence. Feature vectors from the fine-tuning data are extracted via the LLM and clustered. Clusters are enclosed in boxes; a question is flagged as OoD if its feature vector falls outside all boxes. To improve interpretability and robustness, we introduce a regularization loss during fine-tuning that encourages paraphrased questions to stay close in the feature space, and the enlargement of the decision boundary is based on the feature variance within a cluster. Our method complements existing defenses by providing lightweight and interpretable OoD detection.

\end{abstract}

\section{Overview}\label{sec:overview}

Recent developments in Large Language Models (LLMs)~\cite{achiam2023gpt,yang2024qwen2,team2024gemma} have led to impressive performance across a broad spectrum of natural language processing tasks. To align these general-purpose models with the specific demands of downstream applications, fine-tuning techniques such as Low-Rank Adaptation (LoRA)~\cite{hu2022lora} are essential; it enables adaptation to specialized tasks, enforces controlled style and tone, and ultimately enhances contextual relevance and user satisfaction. However, they may produce confident yet incorrect answers when fine-tuned models encounter out-of-distribution (OoD) input, i.e., questions or instructions beyond their intended scope. This behavior poses serious risks, especially in safety-critical applications, where LLMs should ideally abstain from answering outside their fine-tuned boundaries. 

\begin{figure}[t]
\centering
\begin{tikzpicture}
\begin{axis}[
    title={Rejection Rate across Domains},
    ylabel={Rejection Rate (\%)},
    xtick={1,2,3,4,5,6},
    xticklabels={MedQA, Anatomy, Biology, Nutrition, Law, CS},
    xticklabel style={font=\small},
    yticklabel style={font=\large},
    label style={font=\large},
    title style={font=\large},
    legend style={
        font=\scriptsize,
        at={(0.02,0.98)},
        anchor=north west,
        draw=none,
        fill=none,
    },
    ymajorgrids=true,
    width=0.978\columnwidth,
    height=0.8\columnwidth,
    axis x line=bottom,
    axis y line=left,
    tick label style={font=\Large},
    clip=false,
    after end axis/.code={
        \node at (rel axis cs:0.04,-0.15) {\small \textbf{ID}};
        \node at (rel axis cs:0.45,-0.15) {\small \textbf{Near OoD}};
        \node at (rel axis cs:0.90,-0.15) {\small \textbf{Far OoD}};
        \draw[->, thick] (rel axis cs:0.08,-0.15) -- (rel axis cs:0.30,-0.15);
        \draw[->, thick] (rel axis cs:0.60,-0.15) -- (rel axis cs:0.80,-0.15);
    }
]

\addplot[
    color=red,
    mark=*,
    thick
] coordinates {
    (1, 5) (2, 55) (3, 58) (4, 90) (5, 95) (6, 85)
};
\addlegendentry{LoRA-BAM}

\addplot[
    color=blue,
    mark=square*,
    thick
] coordinates {
    (1, 10) (2, 25) (3, 47) (4, 35) (5, 95) (6, 78)
};
\addlegendentry{Mahalanobis Distance}

\addplot[
    color=green!70!black,
    mark=triangle*,
    thick
] coordinates {
    (1, 8) (2, 15) (3, 40) (4, 27) (5, 95) (6, 82)
};
\addlegendentry{Cosine Similarity}

\end{axis}
\end{tikzpicture}
\caption{Domain-wise rejection rate comparison across ID, near-OoD, and far-OoD datasets. Under the same FPR95 setting, LoRA-BAM achieves higher rejection rates on OoD inputs while maintaining low rejection for in-distribution paraphrased MedQA queries.}
\label{fig:result.overview}
\end{figure}
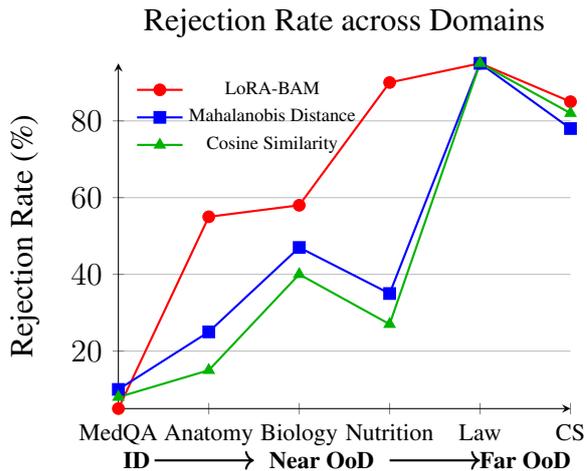

\begin{figure}[t]
\centering

\begin{minipage}{0.55\columnwidth}
\centering
  \includegraphics[width=\textwidth]{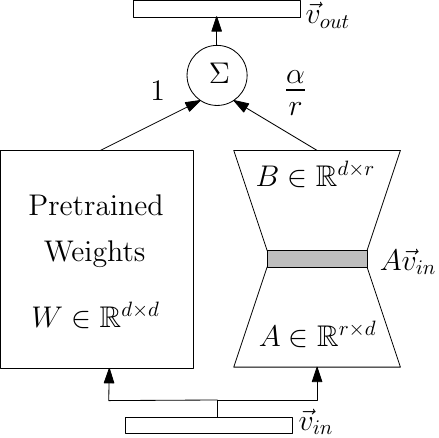}

    \vspace{-1mm}
\caption*{(a)}
    \label{fig:bam.lora}
\end{minipage}
\hfill
\begin{minipage}{0.4\columnwidth}
\centering
\begin{tikzpicture}[scale=1.0]
\draw[->] (2.5,0) -- (3.5,0) node[right] {$\vec{v}_1$};
\draw[->] (2.5,0) -- (2.5,3.5) node[above] {$\vec{v}_2$};

\fill[blue] (3.0, 2.6) circle (0.05);
\fill[blue] (3.2, 2.5) circle (0.05);
\fill[blue] (2.9, 2.4) circle (0.05);
\fill[blue] (3.1, 2.7) circle (0.05);

\fill[blue] (3.0, 1.1) circle (0.05);
\fill[blue] (3.2, 1.0) circle (0.05);
\fill[blue] (2.9, 0.9) circle (0.05);
\fill[blue] (3.1, 1.2) circle (0.05);

\draw[gray] (2.9, 2.4) rectangle (3.2, 2.7);
\draw[gray] (2.9, 0.9) rectangle (3.2, 1.2);

\draw[red, thick] (3.05, 1.8) ellipse [x radius=1.621, y radius=0.239, rotate=89.49];
\end{tikzpicture}
\vspace{-3mm}
\caption*{(b)}

\end{minipage}
\vspace{-2mm}
\caption{LoRA and extracting fracture vectors $A\vec{v}_{in}$ for constructing OoD detectors (a), and an illustrative example indicating the decision boundary between LoRA-BAM (two boxes) and Mahalanobis distance (one ellipse) when enclosing the feature vectors (b)}
\label{fig:two.figure}

\vspace{-5mm}
\end{figure}

In this paper, we propose LoRA-BAM (Boxed Abstraction Monitor for LoRA), a lightweight framework for OoD detection for fine-tuned LLMs utilizing LoRA or its quantized version. Our approach focuses on extracting meaningful representations directly from the LoRA layers (the vector $A\vec{v}_{in}$ in Fig.~\ref{fig:two.figure}a), which are the primary carriers of domain-specific adaptations. OoD detection has been a relatively mature topic in image classification and object detection~\cite{hendrycks2016baseline,lee2018simple,liang2018enhancing,sun2022out,olber2023CVPR,wu2024bam}, and recent work also considers LLMs with built-in capabilities in rejecting OoD inputs utilizing techniques such as Mahalanobis distance~\cite{salimbeni2024beyond} or cosine similarity~\cite{zhang2023outofdistribution,liu2024good}. However, within the recent work in OoD detection of LLMs~\cite{salimbeni2024beyond,liu2024good},  the decision boundary is inherently a convex set (an ellipsoid or spherical cap), making existing methods inherently limited in their detection power. Contrarily, the decision boundary in LoRA-BAM is based on a finite union of boxes (illustrated in Fig.~\ref{fig:two.figure}b), thereby being non-convex in nature.  We not only enclose the vectors using tight boxes, but also properly enlarge the boxes utilizing the variance in the clustered vectors.  

 Finally, to improve robustness against paraphrased questions where we expect the monitor not to reject them, we introduce a regularization loss during fine-tuning that encourages paraphrased questions to stay close in the feature space. This is inspired by recent development of OoD detection techniques in classification or object detection, where the training is adjusted for enabling better OoD detection~\cite{du2022vos,xu2024scaling,lu2024learning,gong2025out,he2025mitigating}. In our case, introducing such a loss function effectively limits the degree of box enlargement, making the method more sensitive to OoD input. 

The main results, as illustrated in Fig.~\ref{fig:result.overview}, indicate that compared to the state-of-the-art method~\cite{salimbeni2024beyond} with their provided benchmark, LoRA-BAM has substantially improved the OoD detection rate.

\section{Monitor Construction Techniques}\label{sec:monitor.construction}

This section explains key components we use in LoRA-BAM to create efficient OoD detectors.

\paragraph{Clustering and Boxed Abstraction} 

Given a LoRA fine-tuned model $f$, let $\mathcal{D}_{train} \defeq \{(q,r)\}$ be the fine-tuning data set of query-response pairs. Given $(q,r) \in \mathcal{D}_{train}$,  let $f^{A}(q) \in \mathbb{R}^{d}$ return the LoRA feature vector of interest following the concept in Fig.~\ref{fig:two.figure}a, and consider $X \defeq \{f^{A}(q) \;|\; (q, r) \in \mathcal{D}_{train}\}$. We apply $k$-means clustering~\cite{sinaga2020unsupervised} to partition the set $X$ into~$m$ clusters:
\[
X = \bigcup_{i=1}^{m} C_i, \quad C_i \cap C_j = \emptyset \text{ for } i \neq j.
\]
For each cluster $C_i \subset \mathbb{R}^k$, we define a bounding box $B_i$ as the axis-aligned box enclosing all points in $C_i$. Specifically, for each dimension $j = 1, \ldots, k$, let
\[
\ell_{i,j} = \min_{\mathbf{x} \in C_i} x_j, \quad u_{i,j} = \max_{\mathbf{x} \in C_i} x_j.
\]
Then the box $B_i$ is defined using Eq.~\eqref{eq:box}, where representing each $B_i$ only requires recording the associated minimum and maximum value in each dimension. 

\begin{equation}\label{eq:box}
    B_i = \left\{ \mathbf{z} \in \mathbb{R}^k \,\middle|\, \ell_{i,j} \leq x_j \leq u_{i,j} \text{ for all } j \right\}
\end{equation}

As $B_i$ has a strict enclosure, it can reject semantically equivalent paraphrased queries. Therefore, it is necessary to enlarge the box, i.e., to replace $\ell_{i,j}$ and $u_{i,j}$ by $\ell_{i,j} - \Delta_{i,j}$ and $u_{i,j} + \Delta_{i,j}$. The decision of $\Delta_{i,j}$ can be done multiple ways, where apart from setting a hard threshold relative to the length $u_{i,j} - \ell_{i,j}$ such as~$1.05$ (i.e., $5\%$ increase), in our implementation, each box $B_i$ is enlarged along every dimension with different ratio. For each dimension $j = 1, \ldots, d$, we compute the standard deviation $\sigma_{i,j}$ of the $j$-th coordinate over all vectors $\vec{z} \in C_i$. Given a hyperparameter $\Delta  > 0$, we expand the box $B_i$ by $\Delta \cdot \sigma_{i,j}$ in both directions along dimension $j$. The enlarged bounds are defined as:
\[
\tilde{\ell}_{i,j} = \ell_{i,j} - \Delta \cdot \sigma_{i,j}, \quad \tilde{u}_{i,j} = u_{i,j} + \Delta \cdot \sigma_{i,j}.
\]
The enlarged box is then given by:
\[
\tilde{B}_i = \left\{ \mathbf{x} \in \mathbb{R}^d \,\middle|\, \tilde{\ell}_{i,j} \leq x_j \leq \tilde{u}_{i,j} \text{ for all } j \right\}
\]

While $\Delta$ is a hyperparameter, to enable fair comparison with other techniques, the criterion of FPR95 (false positive rate at $95\%$ true positive rate) is used. This means that $\Delta$ is adjusted on the ID-only calibration dataset where the OoD filter achieves a~$95\%$ success rate in ``not to consider an ID input as OoD''. Such a technique is commonly used in OoD detection to decide the threshold.

\paragraph{Regularization for Paraphrasing} Although the box construction can lead to extremely tight filtering of OoD samples, it is also highly desirable to ensure that in-distribution samples, such as paraphrased questions, would be accepted. Observe that in the standard fine-tuning process, it is possible that the feature distance vectors $f^{A}(q)$ and $f^{A}(q_p)$ are very distant, thereby causing the monitor to trigger false alarms. 
Therefore, we consider further improving the fine-tuning process to enable better monitorability. Given a query-response pair $(q,r)$, apart from ensuring the correctness of LLM generating~$r$ when inputting~$q$ (normally via cross-entropy loss), we also wish the $q$-rephrased question $q_{p}$ to have a small distance in the LoRA feature space\footnote{In our implementation, we use another LLM to perform paraphrasing of questions.}. 
This is achieved via introducing a new Euclidean Distance Loss $||f^{A}(q), f^{A}(q')||_2$, where pairs of data points are fed into the fine-tuning pipeline. This is analogous to learning domain-invariant representations in a hyperspherical space~\cite{bai2024hypo}.

\paragraph{OoD Query} At inference time, a query~$q'$ is considered as OoD if its transformed representation $f^{A}(q')$ lies outside all the predefined enlarged boxes $\tilde{B_i}$, where checking the box containment amounts to checking if the vector falls inside range of the associated minimum and maximum value, which has total time complexity of~$\mathcal{O}(md)$.

\section{Experiments}\label{sec:experiments}

\subsection{Experimental Setup}

\textbf{Models and Datasets} In our experiments, we use Qwen 2.5 model family~\cite{yang2024qwen2}, where we fine-tune Qwen2.5-0.5B-Instruct on a domain-specific question-answering dataset using LoRA (rank=32) and standard cross-entropy loss using the huggingface peft library\footnote{\url{https://huggingface.co/docs/peft/index}}. To enhance robustness to input variation, we introduced paraphrased versions of the original queries and applied our dual-loss objective, combining cross-entropy with a semantic alignment loss. Specifically, we compute the Euclidean distance between the hidden representations of the original and rephrased inputs extracted from the LoRA-modified projection layer. We follow the experimental protocol introduced in~\cite{salimbeni2024beyond}, where the MedMCQA~\cite{pal2022medmcqa} dataset serves as the ID domain. MedMCQA is a large-scale multiple-choice question dataset focused on medical entrance exams. To construct OoD datasets, we follow the domain structure of the MMLU benchmark~\cite{hendrycks2021measuring}. Specifically, we select Anatomy, Biology, and Nutrition as domains that are semantically related to medicine, forming our near-OoD datasets. In contrast, Law and Computer Science are used as far-OoD datasets, representing subject areas unrelated to the medical domain. This setup allows us to assess the sensitivity of OoD detection methods under varying degrees of distributional shift.

\textbf{OoD detection methods} We evaluate three OoD detection methods. The first is Mahalanobis Distance (MD)~\cite{lee2018simple}, which fits a Gaussian distribution to ID features and uses the distance to this distribution as the OoD score. The second is a cosine similarity–based method~\cite{nguyen2010cosine} that computes the mean ID representation and scores test samples by their cosine similarity to this mean. The third is our proposed method, LoRA-BAM, which was introduced in the previous section.

\begin{table*}[t]
\centering
\scriptsize
\renewcommand{\arraystretch}{1.3}
\setlength{\tabcolsep}{4pt}
\resizebox{1.0\textwidth}{!}{
\begin{tabular}{lcccccc}
\toprule
\multicolumn{7}{c}{\textbf{Model: Qwen2.5-0.5B-Instruct fine-tuned on MedQA Dataset (100 Q-A, denoted as Q(Med))}} \\
\midrule
\textbf{Method} 
& \textbf{Para. In-Distribution} 
& \multicolumn{3}{c}{\textbf{Near OoD}} 
& \multicolumn{2}{c}{\textbf{Far OoD}} \\
\cmidrule(lr){2-2} \cmidrule(lr){3-5} \cmidrule(lr){6-7}
& \makecell[c]{MedQA \\ (Test)} 
& \makecell[c]{Anatomy \\ (100 Q-A)} 
& \makecell[c]{Biology \\ (100 Q-A)} 
& \makecell[c]{Nutrition \\ (100 Q-A)} 
& \makecell[c]{Law \\ (100 Q-A)} 
& \makecell[c]{Computer Science \\ (100 Q-A)} \\
\midrule
LoRA-BAM ($\Delta = 0.2$)              & 64.4\% & 84\%  & 95\%  & 95\%  & 92\%   & 99\% \\
LoRA-BAM ($\Delta = 0.4$)              & 45.4\% & 70\%  & 87\%  & 82\%  & 86\%   & 94\% \\
LoRA-BAM ($\Delta = 0.8$)              & 16\% & 39\%  & 53\%  & 52\%  & 68\%   & 68\% \\
LoRA-BAM ($\Delta = 1$)              & 9\% & 22\%  & 34\%  & 39\%  & 55\%   & 54\% \\
Mahalanobis Distance (TPR = 95 \%)    & 6\%    & 24\%  & 52\%  & 35\%  & 81\%   & 97\% \\
Cosine Similarity (TPR = 95\%)       & 5\%    & 2\%   & 29\%  & 9\%   & 67\%   & 96\% \\
\midrule
\multicolumn{7}{c}{\textbf{Model: Qwen2.5-0.5B-Instruct fine-tuned on paraphrased MedQA Dataset (100 Q-A, denoted as Q*(Med-P)), using new loss}} \\
\midrule
\textbf{Method} 
& \textbf{Para. In-Distribution} 
& \multicolumn{3}{c}{\textbf{Near OoD}} 
& \multicolumn{2}{c}{\textbf{Far OoD}} \\
\cmidrule(lr){2-2} \cmidrule(lr){3-5} \cmidrule(lr){6-7}
& \makecell[c]{MedQA \\ (Test)} 
& \makecell[c]{Anatomy \\ (100 Q-A)} 
& \makecell[c]{Biology \\ (100 Q-A)} 
& \makecell[c]{Nutrition \\ (100 Q-A)} 
& \makecell[c]{Law \\ (100 Q-A)} 
& \makecell[c]{Computer Science \\ (100 Q-A)} \\
\midrule
LoRA-BAM ($\Delta = 0.2$)              & 77\% & 99\%  & 99\%  & 100\%  & 99\%   & 100\% \\
LoRA-BAM ($\Delta = 0.4$)              & 63\% & 99\%  & 98\%  & 100\%  & 99\%   & 100\% \\
LoRA-BAM ($\Delta = 0.8$)              & 48\% & 93\%  & 96\%  & 99\%  & 99\%   & 100\% \\
LoRA-BAM ($\Delta = 1$)              & 40\% & 90\%  & 95\%  & 99\%  & 98\%   & 98\% \\
\textbf{LoRA-BAM (TPR = 95\%)}              & \textbf{3\%} & \textbf{55\%}  & \textbf{58\%}  & \textbf{91\%}  & 95\%   & \textbf{84\%} \\
Mahalanobis Distance (TPR = 95\%)    & 7\%   & 25\%  & 49\%  &35\%  & 96\%   & 77\% \\
Cosine Similarity (TPR = 95\%)       & 9\%    & 17\%   & 41\%  & 26\%  & \textbf{97\%}   & 81\% \\
\bottomrule
\end{tabular}
}
\caption{Comparison of OoD detection methods evaluated on two fine-tuned models: one trained directly on MedQA and the other with paraphrase-aware regularization loss. For the first column (in-distribution), smaller values imply superiority; for the rest, larger values imply superiority. }
\label{tab:double-ood-comparison}
\end{table*}

\subsection{Results}

\textbf{Effectiveness in filtering out pure OoD samples} As shown in Table~\ref{tab:double-ood-comparison}, our proposed LoRA-BAM method consistently achieves superior performance across a range of OoD datasets. On two representative near-OoD domains—Anatomy and Nutrition—LoRA-BAM rejects 55\% and 91\% of OoD samples, significantly outperforming the Mahalanobis-based detector, which rejects only 25\% and 35\% on the same tasks. These results indicate that LoRA-BAM is markedly more sensitive to subtle distributional shifts. On far-OoD datasets, LoRA-BAM remains highly competitive, trailing the best-performing method by only 2\% on the Law domain, while exceeding performance on the other far-OoD case. The performance gains of LoRA-BAM can be attributed in large part to the regularization strategy employed during fine-tuning. When this regularization is removed—i.e., without aligning original and paraphrased samples in the LoRA space—the effectiveness of BAM degrades substantially. As shown in the upper half of Table~\ref{tab:double-ood-comparison}, the unregularized BAM struggles with both near- and far-OoD detection, particularly under stricter margin settings (e.g., $\Delta = 1$), where rejection accuracy drops significantly across all OoD domains. These findings highlight the necessity of our proposed regularization term for shaping a more discriminative and robust representation space, essential for reliable OoD detection.\footnote{In our evaluation, we found that adding the regularization loss (for paraphrased questions) to the existing cross-entropy loss introduces another hyperparameter~$\lambda$ to decide the contribution of the regularization loss. Our initial result with QWen2.5 model showed that $\lambda$ plays a less critical role; we evaluated against $\lambda \in \{0.1,0.5,1,5,7,10\}$ and LoRA-BAM is always better than the other two methods. The result in Fig.~\ref{fig:result.overview} is based on setting $\lambda = 5$.}

\textbf{Robustness against paraphrased ID samples} While existing methods perform well in filtering out pure OoD inputs, they often overlook a critical aspect: whether semantically valid ID variations, such as paraphrased questions, are also inadvertently rejected. Our findings reveal that Mahalanobis distance and cosine similarity baselines reject up to $7\%$ and $9\%$ of paraphrased ID questions, respectively, which can negatively impact user-facing reliability in real-world applications. In contrast, our proposed LoRA-BAM method maintains stronger robustness to such natural ID variations, with a rejection rate as low as 3\%. These results suggest that LoRA-BAM not only achieves strong OoD discrimination but also better preserves coverage over valid but rephrased ID inputs, striking a more favorable balance between selectivity and inclusiveness.

\section{Concluding Remarks}\label{sec:conclusion}

A key strength of LoRA-BAM is its complementarity to other assurance techniques, such as output confidence calibration~\cite{geng2023survey}, ensemble methods~\cite{dietterich2000ensemble}, and retrieval-augmented generation~\cite{gao2023retrieval}. This modularity makes it an attractive choice for deployment in real-world systems requiring multiple layers of defense. We consider future work by introducing dimensionality reduction techniques such as PCA, where, by using geometric structures (i.e., boxes) in a well-defined, low-dimensional space, users and developers can visualize the scope of a fine-tuned model’s capabilities. 

\section*{Limitations}

There remain several limitations in our experiments, and we plan to address these in future work. First, our main experiments are currently limited to Qwen2.5:0.5B, and we intend to scale our experiments to larger models of up to 70B parameters and different model architectures. Next, the fine-tuning dataset is restricted to the one used in prior work~\cite{salimbeni2024beyond} and our newly generated ones, and we plan to introduce additional ones in future work. Finally, we only conduct our experiment with limited random seeds, and more random seeds can be introduced to strengthen the empirical evidence of our results.


\end{document}